\pdfoutput=1

\documentclass[11pt]{article}

\usepackage[preprint]{acl}

\usepackage{times}
\usepackage{latexsym}
\usepackage{siunitx}
\usepackage{footmisc}
\usepackage{footnote}
\usepackage{enumitem}
\usepackage[T1]{fontenc}

\usepackage[utf8]{inputenc}

\usepackage{microtype}

\usepackage{inconsolata}
\usepackage{multirow}

\usepackage{authblk}
\usepackage{graphicx}
\usepackage{subfigure}
\usepackage{booktabs}
\usepackage{CJK}
\usepackage{hyperref}
\usepackage{colortbl}
\definecolor{mygray}{gray}{.9}
\title{LegalAgentBench: Evaluating LLM Agents in Legal Domain}

\author{
\textbf{Haitao Li$^{1}$\thanks{\ Equal contributions.}\ \thanks{Work done when HT and JJ interned at Zhipu AI.} , Junjie Chen$^{1}$\footnote[1]\footnote[2], Jingli Yang$^{2}$, Qingyao Ai$^{1}$\thanks{Corresponding author}, Jia Wei$^{2}$, Youfeng Liu$^{2}$, Kai Lin$^{3}$} \\
\textbf{Yueyue Wu$^{1}$\footnote[3], Guozhi Yuan$^{4}$, Yiran Hu$^{5}$, Wuyue Wang$^{6}$, Yiqun Liu$^{1}$, Minlie Huang$^{1}$} \\
$^1$Department of Computer Science and Technology, Tsinghua University, $^2$Zhipu AI, \\
$^3$Shanghai Amarsoft Enterprise Credit Information Service Co.,Ltd,\\
$^4$Central South University,$^5$University of Waterloo, $^6$University of Notre Dame\\
\texttt{liht22@mails.tsinghua.edu.cn}
}

\begin{document}
\maketitle
\begin{abstract}
With the increasing intelligence and autonomy of LLM agents, their potential applications in the legal domain are becoming increasingly apparent.
However, existing general-domain benchmarks cannot fully capture the complexity and subtle nuances of real-world judicial cognition and decision-making.
Therefore, we propose \textbf{LegalAgentBench}, a comprehensive benchmark specifically designed to evaluate LLM Agents in the Chinese legal domain.
LegalAgentBench includes 17 corpora from real-world legal scenarios and provides 37 tools for interacting with external knowledge. 
We designed a scalable task construction framework and carefully annotated 300 tasks. These tasks span various types, including multi-hop reasoning and writing, and range across different difficulty levels, effectively reflecting the complexity of real-world legal scenarios.
Moreover, beyond evaluating final success, LegalAgentBench incorporates keyword analysis during intermediate processes to calculate progress rates, enabling more fine-grained evaluation.
We evaluated eight popular LLMs, highlighting the strengths, limitations, and potential areas for improvement of existing models and methods.
LegalAgentBench sets a new benchmark for the practical application of LLMs in the legal domain, with its code and data available at \url{https://github.com/CSHaitao/LegalAgentBench}.

\end{abstract}

\section{Introduction}
Recent advances in large language models (LLMs) have significantly increased the field of artificial intelligence~\cite{achiam2023gpt,bai2023qwen,touvron2023llama}.  With their expansive neural networks and vast training datasets, LLMs have made remarkable strides in various natural language processing tasks, such as text generation and machine translation~\cite{li2024llms,li2024calibraeval,zhao2024surveylargelanguagemodels}.
At the same time, the rapid evolution of LLMs is transforming the traditional legal industry, empowering legal professionals to handle tasks such as legal research, contract drafting, and case analysis with greater efficiency.
As a result, LLMs are rapidly becoming indispensable tools in modern legal workflows~\cite{cui2023chatlaw,li2024lexeval,guha2024legalbench,li2023sailer}.

Despite their immense potential, LLMs still face challenges in tackling complex legal issues, as real-world legal tasks often require multistep reasoning and specialized expertise beyond their current capabilities. A promising solution lies in the development of LLM-as-Agent systems~\cite{dorri2018multi,liu2023agentbench}. These agents can engage in step-by-step reasoning and acquire specialized knowledge through iterative interactions with external tools. Their impressive capabilities have attracted significant interest from both academia and industry~\cite{qin2023toolllm,li2024bladeenhancingblackboxlarge}.

Although LLM agents show great promise, the lack of standardized benchmarks to evaluate their performance in legal scenarios is a major challenge.
Existing frameworks, such as AgentBench~\cite{liu2023agentbench} and ToolBench~\cite{qin2023toolllm}, are effective in assessing LLM agents in general domains. However, the insights gained from these evaluations often have limited relevance in highly specialized fields such as the legal domain~\cite{li2024lexeval}.
Moreover, existing datasets in the legal domain often focus on relatively basic tasks, such as legal case retrieval~\cite{li2024lecardv2,li2024delta} or judgment prediction~\cite{li2024lexeval}. In contrast, legal practice is significantly more complex, involving in-depth case analysis, legal reasoning, and comprehensive judgments based on a vast body of laws and precedents. Current datasets and evaluation systems fall short of thoroughly testing these advanced, multidimensional legal capabilities.

To fill this gap, we developed \textbf{LegalAgentBench}, a comprehensive benchmark designed to evaluate the capabilities of LLM agents in the Chinese legal domain.
Based on real-world legal needs, it includes 17 specialized corpora and 37 tools to interact with external knowledge. Within this framework, LLM agents coordinate and utilize these tools to address specific legal tasks.

LegalAgentBench is highlighted in the following three aspects:

\begin{itemize}[leftmargin=*]
    \item \textbf{Focus on Authentic Legal Scenarios:} LegalAgentBench is the first dataset to evaluate LLM agents in legal scenarios. It requires LLMs to demonstrate a solid understanding of legal principles, enabling them to appropriately select and utilize tools to solve complex legal problems. This represents a significant step forward in advancing the application of LLM agents in legal scenarios.

    \item \textbf{Diverse Task Types and Difficulty Levels:} 
    LegalAgentBench adopts a scalable task construction framework aimed at comprehensively covering various task types and difficulty levels. Specifically, we construct a planning tree based on the dependencies between the corpus and tools, and select tasks through hierarchical sampling and a maximum coverage strategy. 
    Finally, we constructed 300 distinct tasks, including multi-hop reasoning and writing tasks, to comprehensively evaluate the LLM's capabilities.

    \item \textbf{Fine-Grained Evaluation Metrics:} Rather than relying solely on final success rates as evaluation criteria, LegalAgentBench introduces the process rate through the annotation of intermediate steps, enabling fine-grained evaluation. This approach provides deeper insights into an agent's capabilities and identifies areas for improvement beyond the final result.
\end{itemize}

We evaluated a variety of commercial and open-source LLM agents, identifying several potential weaknesses in their current capabilities. These observations provide valuable insights, inspire new ideas, and pinpoint areas for further research and improvement.

\section{Related Work}

\subsection{LLM Agents}

Recently, large language models (LLMs) have achieved significant advancements in their application as intelligent agents. Leveraging the capabilities in natural language understanding and generation, LLM agents can efficiently solve complex tasks by appropriately invoking external tools such as calculators, search engines, and domain-specific APIs~\cite{yaoreact,Wang_2024,wang2023plan,shinn2024reflexion,chu2024pre}.

When faced with complex tasks, LLM agents need to devise appropriate plans and strategies to ensure efficient execution. Chain-of-Thought (CoT)~\cite{wei2022chain} reasoning, a pioneering technique in this domain, enhances the reasoning capabilities of agents by decomposing challenging reasoning tasks into smaller, more manageable steps. Building upon this foundation, a series of advanced reasoning strategies have emerged, offering new perspectives for task planning and problem-solving. For instance, ReAct~\cite{yaoreact} decouples reasoning and action, alternating between thinking steps and action steps, thereby significantly improving the planning efficiency for complex tasks.

Furthermore, the capabilities of LLM agents are further enhanced through deep integration with external tools. Methods such as HuggingGPT~\cite{shen2024hugginggpt} exemplify this approach by positioning LLMs as controllers that decompose complex tasks into subtasks, invoke appropriate specialized models, and integrate the results to produce a final response. The LLM+P~\cite{liu2023llm} approach combines LLMs with a symbolic planner based on the Planning Domain Definition Language (PDDL), leveraging LLMs to formulate problems in PDDL syntax and employing a planning solver to generate solutions.

\subsection{Benchmarks on LLM Agents}

With the rapid advancement of LLM agents, the demand for evaluating their reasoning and decision-making capabilities in complex tasks has grown significantly. Numerous benchmarks have been developed, providing essential references for the systematic study of LLM agents~\cite{liu2023agentbench,zhuang2023toolqa,ye2024tooleyes,yao2024tree,huang2023metatool,10.1145/3539618.3591874}.

AgentBench~\cite{liu2023agentbench} is a multidimensional benchmark platform that spans eight distinct environments, including operating systems, databases, and knowledge graphs. It systematically evaluates the performance of LLM agents in multi-turn open-ended generation settings.
AgentBoard~\cite{ma2024agentboard} focuses on the systematic evaluation of LLMs in multi-turn interactions. It introduces fine-grained progress rate metrics and multidimensional analysis tools to comprehensively assess LLMs' multitasking capabilities in partially observable environments, offering a scientific framework for evaluating interactive LLM performance.
ToolQA~\cite{zhuang2023toolqa} covers eight domains and 13 tools, with tasks specifically designed to require external tools and reference materials for problem-solving. This avoids reliance solely on the model's internal knowledge, establishing a new benchmark for assessing LLMs' tool-using abilities.
T-Eval~\cite{chen2024t} decomposes the tool utilization abilities of LLMs into six dimensions: planning, reasoning, retrieval, understanding, instruction execution, and result evaluation. This framework comprehensively measures LLMs' performance in using external tools across multiple levels.

These benchmarks provide diverse tools for exploring the capability boundaries of LLM agents. However, they primarily focus on evaluating LLMs in general domains, lacking systematic guidance and targeted frameworks for vertical domains such as law and medicine. To address this gap, this paper introduces a dataset specifically designed to evaluate LLM agents in the legal domain, offering valuable insights and support for research and practice.
\section{LegalAgentBench}


\subsection{Overview}

LegalAgentBench consists of three key components: the environment, tools, and tasks. LLM agents rely on these tools to interact with the environment and tackle specific tasks. We define the environment as a text-based corpora, where both the observation and action spaces are represented in natural language.
Specifically, the environment consists of 17 distinct corpora, 14 of which are tabular databases for lookup, while 3 are document collections for retrieval. 
To facilitate interaction with the environment, we provide 37 specialized tools for tasks such as text retrieval, database operations, and mathematical computations.

\begin{figure}[t]
\vspace{-5mm}
\includegraphics[width=\linewidth]{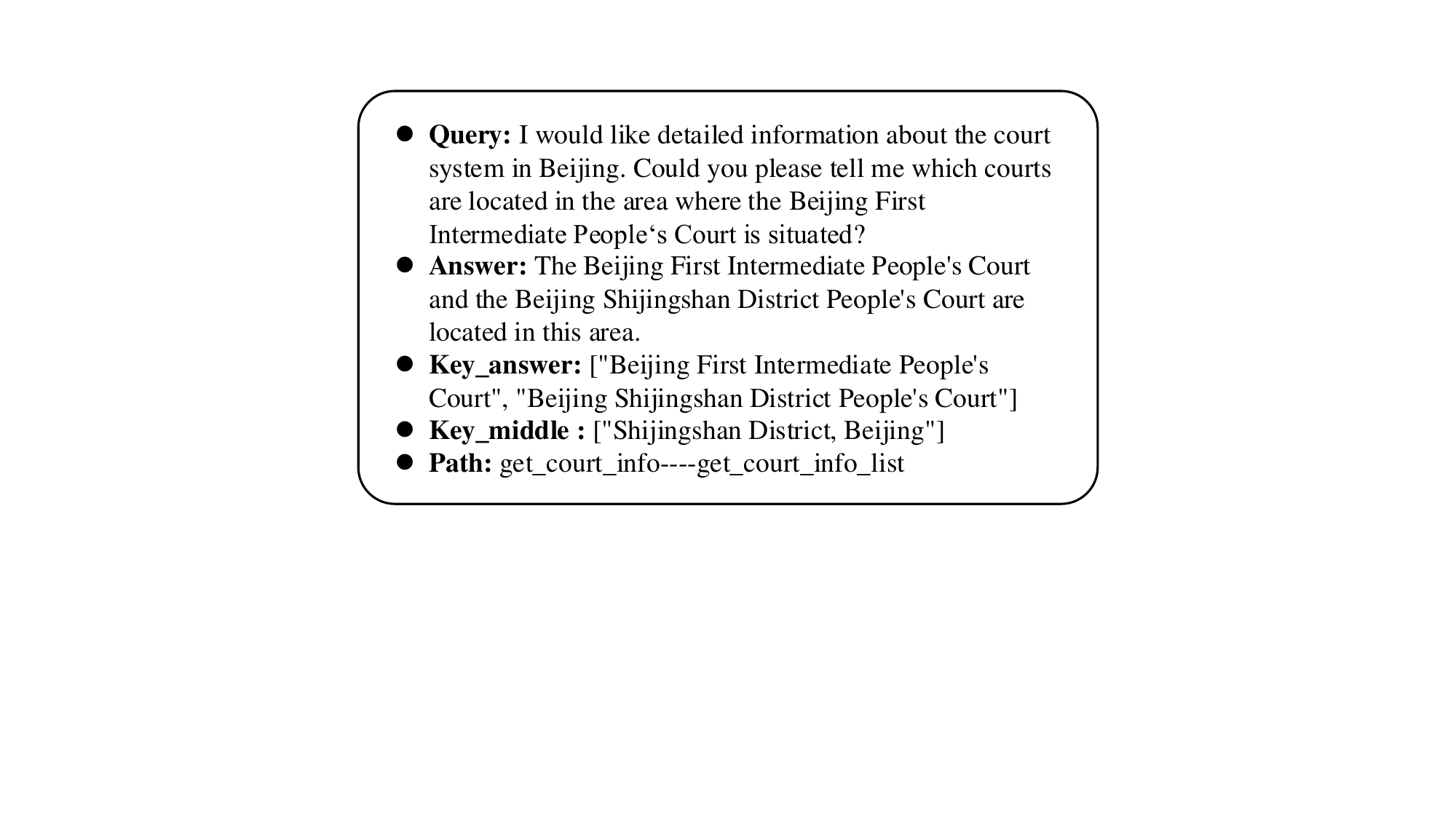}
\vspace{-5mm}
\caption{A task example in LegalAgentBench (translated from Chinese).}
\label{figure:example}
\vspace{-5mm}
\end{figure}

In addition, LegalAgentBench includes 300 tasks of varying difficulty levels and types. Figure~\ref{figure:example} illustrates a task example. The key\_answer represents the keywords in the answer, used to evaluate the success rate of the final result. The key\_middle refers to the keywords in the intermediate steps of solving the task, providing a more granular evaluation. These keywords are derived from the observations returned by successful tool calls. The Path refers to the correct solution path for solving the task. A longer path signifies a higher level of difficulty associated with the task.

Upon receiving the task, the LLM agent gradually selects the appropriate tools and obtains corresponding feedback. After receiving feedback, the LLM agent updates its internal state based on the observation and then determines the next action. The agent's actions typically involve selecting the appropriate tool and inputting the necessary parameters, or adjusting the query strategy based on the feedback. Each action generates a new observation, and this cycle continues until the agent completes the task or reaches the predefined termination condition.

\subsection{Corpora and Tools}

In LegalAgentBench, we developed 17 real-world datasets, including 14 tabular databases designed for lookups and 3 document repositories intended for retrieving relevant information. Table~\ref{table:corpus} presents the basic information and a brief overview of these corpora. More details can be found in Appendix~\ref{sec:Corpus}.

Specifically, we gather publicly available real-world data, encompassing information about courts, law firms, listed companies, and their associated legal cases. Additionally, we compile legal knowledge, articles, and guiding cases to create comprehensive and searchable corpora. These corpora are sourced from real-world scenarios and can evolve over time, mitigating the risk of LLM overfitting. Due to the sensitivity of the legal domain, we discuss the license and ethical considerations in Appendix \ref{sec:Discussion}.

\begin{table*}[t]

\centering
\small
\vspace{-3mm}
\begin{tabular}{cllrc}
\hline
ID & \multicolumn{1}{l}{Corpus} & \multicolumn{1}{c}{Format} & Size   & \multicolumn{1}{c}{Brief overview}                           \\ \hline
1  & CompanyInfo              & Tabular Database           & 695    & Basic information of listed companies                           \\
2  & CompanyRegister          & Tabular Database           & 10,125 & Registration information of listed companies                    \\
3  & SubCompanyInfo           & Tabular Database           & 9,433  & Investment information of listed companies                      \\
4  & LegalDoc                 & Tabular Database           & 24,372 & Legal cases involving listed companies                          \\
5  & LegalAbstract            & Tabular Database           & 1,200  & Summary information of legal cases                              \\
6  & CourtInfo                & Tabular Database           & 3,413  & Basic information of courts                                     \\
7  & CourtCode                & Tabular Database           & 3,348  & Levels and administrative division codes of courts              \\
8  & LawfirmInfo              & Tabular Database           & 4,768  & Basic information of law firms                                  \\
9  & LawfirmLog               & Tabular Database           & 101    & Service information of law firms                                \\
10 & AddrInfo                 & Tabular Database           & 19,533 & Province, city, and district corresponding to the address       \\
11 & RestrictionCase          & Tabular Database           & 46     & Cases involving restrictions on high consumption \\
12 & FinalizedCase               & Tabular Database           & 119    & Cases closed upon final enforcement              \\
13 & DishonestyCase           & Tabular Database           & 13     & Cases involving dishonest judgment debtors       \\
14 & AdministrativeCase       & Tabular Database           & 443    & Cases involving administrative penalty    \\
15 & LegalKonwledge           & Retrieval Corpus           & 26,951 & Knowledge from legal books                                      \\
16 & LegalArticle             & Retrieval Corpus           & 55,347 & Legislatively enacted legal articles                            \\
17 & LegalCases               & Retrieval Corpus           & 2,370  & Officially published guiding cases                              \\ \hline
\end{tabular}
\caption{Basic information of the corpora.}
\label{table:corpus}
\vspace{-5mm}
\end{table*}

To obtain information from these reference corpora, we designed 37 tools that are available to the LLM Agents. These tools are primarily divided into the following categories:

\begin{itemize}[leftmargin=*]

    \item \textbf{Text Retrievers:} These tools are responsible for retrieving content relevant to a given query from document repositories. We use Embedding-3~\footnote{\url{https://bigmodel.cn/}} as the default retriever. Additionally, users can integrate more advanced retrieval models to further improve query results. LegalAgentBench includes three text retrievers.  For each retriever, the LLM can use the \textit{Number} parameter to specify the number of relevant documents to be returned.

    \item \textbf{Mathematical Tools:} These tools perform basic arithmetic operations such as addition, subtraction, multiplication, and division. Additionally, they can handle more complex tasks like sorting data, and computing maximum or minimum values. There are five mathematical tools in LegalAgentBench. 

    \item \textbf{Database Tools:} These tools interact with specific databases to extract content from particular columns based on predefined queries. They allow the LLM agents to access structured data, providing tailored responses that match the criteria defined in the query. There are 28 database tools in LegalAgentBench.  For each Database Tool, the LLM can use the parameter \textit{Column} to control the attributes returned from the corresponding tabular database.

    \item \textbf{System Tools:} The System tool currently only includes \textit{Finish}, which parses the execution feedback and returns the answer to complete the task.

\end{itemize}

Due to space constraints, detailed descriptions of each tool, including their inputs, outputs, and usage examples, are provided in Appendix~\ref{sec:Tool}.

\begin{figure*}[t]
\centering
\vspace{-5mm}
\includegraphics[width=\linewidth]{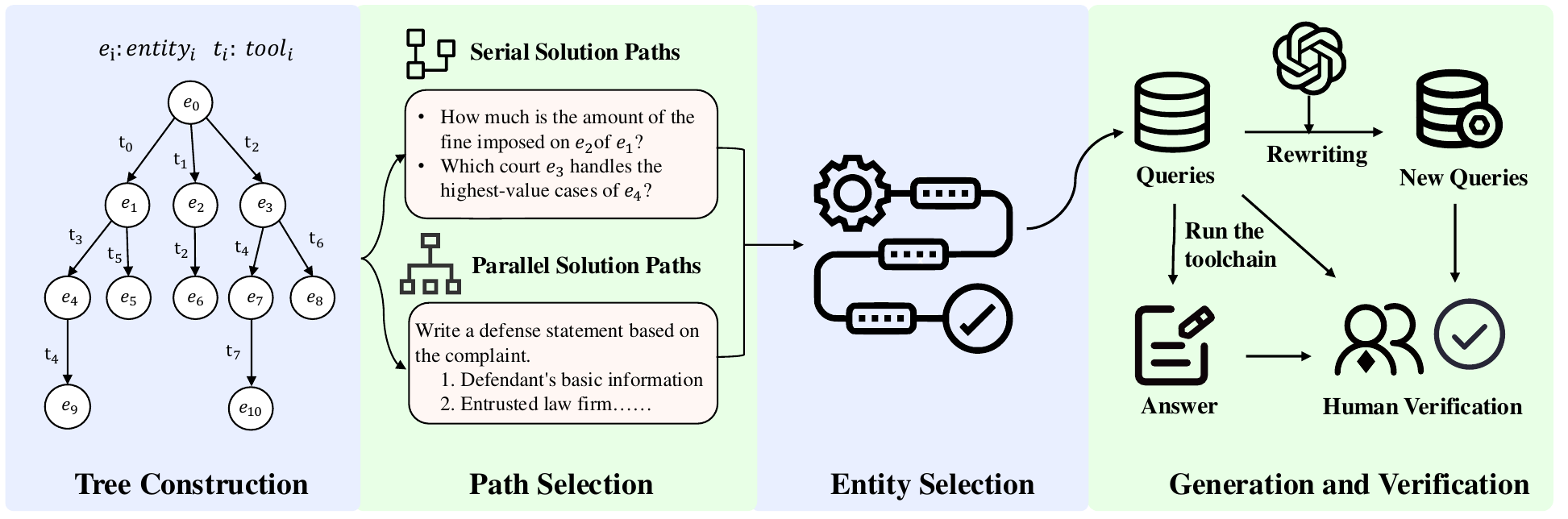}
\vspace{-3mm}
\caption{The overview of the task construction process in LegalAgentBench.}
\label{figure:task}
\vspace{-5mm}
\end{figure*}

\subsection{Tasks}
\subsubsection{Task Definition}
LegalAgentBench employs a unified framework to standardize all tasks, offering a formalized definition of the agent's interaction process in legal scenarios.
At each time step $t$, the agent performs an action $a_t$, receives an observation $o_t$, and updates its current state $s_t$ accordingly. This iterative process continues until the task is successfully completed or the maximum iteration limit $T$ is reached. 

$s_t$ represents the agent's perceptual state at time $t$, encompassing the acquired contextual information and the environmental state. The state $s_t$ is updated according to the following formula:
$$
s_{t+1} = u(s_t,a_t,o_t),
$$
where $u$ is the state update function, responsible for integrating the agent's actions and feedback into a new perceptual state.

$a_t$ represents the specific action taken by the agent in the state $s_t$, including tool invocation or information generation. The action $a_t$ is determined by the state $s_t$ and the sequence of past observations.
$$
a_t = \pi(s_t,{o_1,o_2,...,o_{t-1}}),
$$
where $\pi$ is the agent's decision-making policy.

$o_t$ is the feedback information received by the agent after invoking a tool, including successful results or error messages.
$$
o_t = f(a_t)
$$
Here, $f$ is the feedback function. If $a_t$ is a valid action, it returns the result of the operation; otherwise, it returns an error message.

\subsubsection{Task Construction}
During the task generation phase, our primary goal is to leverage corpus and tools to construct a diverse set of questions. The core principle is to ensure that tasks encompass a wide range of difficulty levels and types, enabling a comprehensive evaluation. In this section, we propose a scalable task construction process that can be extended to incorporate new knowledge bases and tools. The entire process is illustrated in Figure \ref{figure:task}.

\begin{table*}[t]
\vspace{-5mm}
\centering
{%
\begin{tabular}{cccccccc}
\hline
Attribute                  & 1-hop & 2-hop & 3-hop & 4-hop  & 5-hop  & Writing & ALL    \\ \hline
\# Task & 80 & 80 & 60 & 40 & 20 & 20 & 300 \\

Avg. length per query   & 88.29 & 87.90 & 99.37 & 118.33 & 110.25 & 1059.95 & 160.65 \\
Avg. length per answer     & 74.20 & 40.84 & 45.53 & 63.48  & 86.20  & 678.75  & 99.24  \\
\# Avg. key\_answer per query & 1.88  & 1.44  & 1.20  & 1.40   & 2.25   & 10.25   & 2.14   \\
Avg. length of key\_answer & 10.59 & 5.94 & 6.07 &6.59 &6.93 &12.58 &9.28 \\
\# Avg. key\_middle per query & 0.13  & 1.45  & 2.87  & 4.78   & 5.60   & 6.20    & 2.42   \\
Avg. length of key\_middle     & 9.20 &9.72 &10.95 &11.35 &11.25 &7.21 &10.24   \\
 \hline
\end{tabular}
}
\vspace{-3mm}
\caption{Detailed Statistics of LegalAgentBench Tasks.}
\label{table:statistics}
\vspace{-5mm}
\end{table*}

\textbf{Planning Tree Construction:} 
To better organize and structure the task, we first construct a planning tree based on the call relationships between the available tools. The root node of the tree represents the unknown entity, which typically corresponds to the starting point of the task. Each branch of the tree corresponds to a tool that can be utilized by the entity, while the child nodes contain the information obtained after invoking these tools. For each child node, subsequent tools may be called progressively, forming deeper branches. In this planning tree, each path represents a potential solution pathway, and the information at the leaf nodes corresponds to the final requirements of the task.

\textbf{Path Selection:} 
We select different paths from the planning tree to construct tasks that cover a variety of types and difficulty levels. 
In the planning tree, the depth of each branch represents the complexity of the task solution, while the breadth of all branches reflects the diversity of task types.
Therefore, we extract solution paths layer by layer to ensure coverage of different difficulty levels. Moreover, we minimize overlap between solution paths for tasks of different difficulty levels, expanding the coverage of all branches to encompass a broader range of task types. We ultimately developed serial solution paths ranging from 1-hop to 5-hop.

In addition to using tools sequentially, our solution paths also incorporate the parallel use of tools. 
We introduce the task of writing a defense document as a typical example.
In this scenario, the LLM must query basic information about the plaintiff, defendant, and their lawyer based on a template, while simultaneously retrieving fundamental legal knowledge and relevant articles to formulate a defense against the complaint. In Appendix \ref{sec:Task}, we provide examples for better understanding.

\textbf{Entity Selection:} 
After determining the solution path, we select entities to formulate the complete question.
It is important to note that not all initial entities can complete the intended solution path, as there may be cases where the tool call returns no result.  To address this issue, we iterate through all possible entities and select two successfully executed entities for each path.

\textbf{Question Rewriting:} 
After the above steps, we obtain the appropriate entities and can automatically generate multi-hop questions such as ``What is $entity_3$ of $entity_2$ of $entity_1$?''. However, this type of question not only deviates from actual usage habits but also directly exposes the solution path to the model. To better align with real-world scenarios and obscure the solution path, we rephrase the questions using GPT-4 to make them more flexible and in line with human needs. The specific prompt can be found in Appendix \ref{sec:Task}.

\textbf{Answers Generation:}
For each question, since the entities and the toolchain used in the solution path are known, we can programmatically extract the answer from the reference corpora through the appropriate parameterized toolchain. This enables the automatic generation of correct answers, even for complex multi-hop reasoning tasks.

\textbf{Human Verification:} We manually validate all questions, solution paths, and answers in the tasks. This involves revising incompatible queries and tools, refining questions that deviate from everyday usage, and addressing queries related to specialized tools. Detailed guidelines for the validation process are provided in Appendix \ref{sec:Guidelines}.

\subsubsection{Task Evaluation}

Given the high accuracy requirements in the legal domain, we primarily evaluate performance using keyword matching to calculate the success rate.
Specifically, we extract keywords from tool call results and record them as key\_answer. The overlap between the agent's output and these keywords serves as a measure of its capabilities.

However, relying solely on the success rate fails to capture subtle differences, as it cannot distinguish between tasks that are nearly completed and those that are unfinished. To address this issue, we provide keywords not only for the final answers but also for the intermediate solution steps. Using these keywords, we calculate the progress rate, enabling a more fine-grained evaluation of the agent's performance at various stages of task completion.
We provide the detailed calculation for the success rate and progress rate in Appendix~\ref{sec:Implementation}.

\subsubsection{Task Statistics}
After careful human verification, LegalAgentBench includes a total of 300 tasks across 6 different types. Table \ref{table:statistics} presents the detailed statistics of the tasks. Overall, LegalAgentBench has a well-balanced distribution of difficulty and task types, making it an effective tool for evaluating the capabilities of agents in the legal domain.


\section{Experiment}

\subsection{Experiment Setup}

\begin{table*}[]
\centering
\resizebox{\textwidth}{!}{%
\begin{tabular}{llcccccccr}
\hline
Model                              & Method & 1-hop           & 2-hop           & 3-hop           & 4-hop           & 5-hop           & Writing         & ALL             & Tokens   \\ \hline
\multirow{3}{*}{GLM-4}              & P-S    & 0.7588          & 0.4229          & 0.1806          & 0.3708          & 0.1750          & 0.5778          & 0.4509          & 5,100,468  \\
                                   & P-E    & 0.7838          & 0.4042          & 0.2056          & 0.3083          & 0.1600          & 0.5469          & 0.4461          & 9,849,924  \\
                                   & ReAct  & 0.8787          & 0.6771          & 0.4167          & 0.3875          & 0.2433          & 0.5937          & 0.6057          & 11,920,863 \\ \hline
\multirow{3}{*}{GLM-4-Plus}         & P-S    & 0.8519          & 0.4667          & 0.4167          & 0.3583          & 0.1167          & 0.7522          & 0.5406          & 5,657,827  \\
                                   & P-E    & 0.8419          & 0.5000          & 0.3667          & 0.3458          & 0.1167          & 0.7679          & 0.5363          & 9,422,692  \\
                                   & ReAct  & 0.9131          & 0.8104          & 0.6417          & 0.6167          & 0.4300          & 0.7659          & 0.7499          & 11,739,861 \\ \hline
\multirow{3}{*}{LLaMa3.1-8B} & P-S    & 0.3406          & 0.1333          & 0.0333          & 0.0375          & 0.1083          & 0.2382          & 0.1612          & 9,279,701  \\
                                   & P-E    & 0.3510          & 0.1042          & 0.0250          & 0.0500          & 0.0667          & 0.3484          & 0.1607          & 13,649,741 \\
                                   & ReAct  & 0.6019          & 0.1542          & 0.0750          & 0.0708          & 0.0600          & 0.0872          & 0.2359          & 50,661,127 \\ \hline
\multirow{3}{*}{Qwen-max}          & P-S    & 0.8469          & 0.4958          & 0.4083          & 0.3792          & 0.2333          & 0.4836          & 0.5381          & 4,800,345  \\
                                   & P-E    & 0.8594          & 0.5896          & 0.3583          & 0.4083          & 0.3017          & 0.5539          & 0.5695          & 9,884,307  \\
                                   & ReAct  & 0.9062          & 0.7917          & 0.6333          & 0.5833          & 0.6083          & 0.6662          & 0.7422          & 11,473,873 \\ \hline
\multirow{3}{*}{Claude-sonnet}     & P-S    & 0.8137          & 0.6354          & 0.4833          & 0.3750          & 0.2400          & 0.8395          & 0.6051          & 7,100,962  \\
                                   & P-E    & 0.8700          & 0.6771          & 0.5333          & 0.4667          & 0.4717          & 0.8610          & 0.6703          & 13,566,119 \\
                                   & ReAct  & 0.8950          & 0.6979          & 0.4750          & 0.4792          & 0.4567          & 0.6567          & 0.6579          & 32,878,858 \\ \hline
\multirow{3}{*}{GPT-3.5}           & P-S    & 0.4906          & 0.2396          & 0.0500          & 0.1000          & 0.0667          & 0.0333          & 0.2247          & 5,007,391  \\
                                   & P-E    & 0.4906          & 0.2062          & 0.0667          & 0.0625          & 0.0500          & 0.0405          & 0.2135          & 9,597,807  \\
                                   & ReAct  & 0.6421          & 0.2854          & 0.1167          & 0.1000          & 0.1333          & 0.0852          & 0.2986          & 11,357,664 \\ \hline
\multirow{3}{*}{GPT-4o-mini}       & P-S    & 0.7117          & 0.3375          & 0.2750          & 0.2583          & 0.1250          & 0.6314          & 0.4196          & 5,482,556  \\
                                   & P-E    & 0.7444          & 0.3771          & 0.3250          & 0.2417          & 0.1417          & 0.6681          & 0.4503          & 10,861,492 \\
                                   & ReAct  & \textbf{0.9333}          & 0.6500          & 0.4000          & 0.4208          & 0.2583          & 0.6087          & 0.6161          & 13,332,418 \\ \hline
\multirow{3}{*}{GPT-4o}             & P-S    & 0.7856          & 0.4437          & 0.2750          & 0.1875          & 0.2250          & 0.7971          & 0.4760          & 4,153,333  \\
                                   & P-E    & 0.8106          & 0.3979          & 0.3167          & 0.2167          & 0.2767          & \textbf{0.8643} & 0.4906          & 7,261,238  \\
                                   & ReAct  & 0.9262 & \textbf{0.8396} & \textbf{0.7500} & \textbf{0.6417} & \textbf{0.6117} & 0.6541          & \textbf{0.7908} & 11,206,957 \\ \hline
\end{tabular}%
}
\caption{The success rate of different baselines on LegalAgentBench. P-S represents the Plan-and-Solve method, and P-E represents the Plan-and-Execute method. The best results are highlighted in bold.}
\label{table:result}
\vspace{-5mm}
\end{table*}

\subsubsection{Baselines}
We evaluated eight well-known LLMs on LegalAgentBench: GLM-4~\cite{glm2024chatglm}, GLM-4-Plus~\cite{glm2024chatglm}, LLaMA3.1-8B-instruct~\cite{touvron2023llama}, Qwen-max~\cite{bai2023qwen}, Claude-sonnet (claude-3-5-sonnet-20241022), GPT-3.5 (gpt-3.5-turbo-1106), GPT-4o-mini (gpt-4o-mini-2024-07-18), and GPT-4o (gpt-4o-2024-08-06). 
Except for LLaMA3.1-8B-Instruct, other LLMs are evaluated through API calls.
To ensure the reproducibility of the results, we set the temperature for all LLMs to 0.

For each LLM, we implemented three different methods. 1) \textbf{Plan-and-Solve}~\cite{wang2023plan}: Outline a complete plan and execute it step by step. 2) \textbf{Plan-and-Execute}~\cite{topsakal2023creating}: Develop a multi-step plan and complete it sequentially. After completing a task, the LLM can reassess the plan and make appropriate adjustments. 3) \textbf{ReAct}~\cite{yaoreact}: Perform reasoning incrementally through the ``thought-action-observation'' process, integrating reasoning and tool usage.

When given a task, the model first determines which tools are needed, and then uses the selected tools to gradually solve the task. 
When the LLM outputs \textit{Finish} or reaches the maximum iteration limit $T=10$, it summarizes the current trajectory and provides the final answer. We included three examples for each process to guide the model in using the tools and following the specified output format. Additional implementation details are available in Appendix~\ref{sec:Implementation}.

\subsubsection{Metrics}
We apply three evaluation metrics, \textbf{Success Rate}, \textbf{Process Rate}, and \textbf{BERT-Score}~\cite{zhang2019bertscore}, to comprehensively evaluate the performance. Specifically, the success rate calculates the proportion of key\_answer elements included in the LLM's answer. The process rate further incorporates key\_middle, measuring the ratio of key\_middle and key\_answer elements present in the answer.
Moreover, BERTScore is applied to compute the textual similarity between the generated answer and the reference answer, thus assessing the quality and accuracy of the output.
In addition to these metrics, we also report the number of tokens consumed by the LLMs as a reference for efficiency.
Due to space limitations, we report only the success rate in the main result. More results can be found in the Appendix \ref{sec:Evaluation}.

\subsection{Main Results}

The performance comparison between different LLMs and baselines on LegalAgentBench is shown in Table \ref{table:result}. More experimental results can be found in Appendix \ref{sec:Evaluation}. Based on the experimental results, we draw the following observations.

\begin{itemize}[leftmargin=*]
\item \textbf{Comparing Different LLMs.}
In experiments across different LLMs, GPT-3.5 and LLaMA3.1-8B demonstrated poor performance, with success rates on LegalAgentBench below 30\%. This may stem from their limited ability to effectively use tools, restricting their problem-solving capacity in complex legal tasks.
GLM4, GLM4-Plus, Qwen-Max, and GPT-4o-mini consumed tokens at similar levels. However, GLM4-Plus and Qwen-Max demonstrated superior performance.
Claude-Sonnet also achieved competitive results, delivering the best performance under both P-S and P-E methods. However, it often required more tokens compared to other LLMs. Under the ReAct method, GPT-4o achieved the best performance with relatively fewer tokens, reaching a success rate of 79.08\%.
Overall, LegalAgentBench effectively differentiates the capabilities of various LLMs, with those demonstrating stronger tool usage and logical reasoning achieving superior performance.

\item \textbf{Comparing different methods.}
By comparing different methods, we found that ReAct typically produces better results for multi-hop questions. However, this advantage comes with higher token consumption, suggesting that allowing more time for reasoning could improve performance.
Additionally, we found that when LLMs are limited in capability, the P-E method doesn't always outperform P-S. LLMs like GLM-4, LLaMa3.1-8B, and GPT-3.5 show similar trends. This may be due to plan updates increasing the context length, which reduces the effectiveness of the attention mechanism.
The performance gap between different reasoning methods for the same LLM can reach 65\%, emphasizing that effective methods better utilize the LLM's potential. Additionally, when designing effective reasoning methods, it is also crucial to balance model capability, reasoning time, and token consumption.

\item \textbf{Comparing different types of queries.}
As shown in Table~\ref{table:result}, for multi-hop questions, all baselines showed decreased performance as the number of hops increased.
The best performance achieved 93\% success on the 1-hop question, but only 61\% on the 5-hop question. This indicates that the questions in LegalAgentBench cover a wide range of difficulty levels.
For the Writing task, we observed that ReAct performed poorly compared to other methods.
We believe this may be due to the step-by-step update mechanism, which repeatedly attempts to solve individual steps when errors occur.
In tasks like writing, which require parallel processing, this approach may overlook other potential paths that could lead to other answers. This also highlights that the diverse types of questions in LegalAgentBench effectively assess the potential of different methods.

\end{itemize}

\subsection{Analysis}
In this section, we analyze the unique challenges faced by LLM agents in LegalAgentBench and the potential improvements.

\textbf{Lack of specialized legal knowledge.}
The terminology and concepts in the legal domain are highly specialized, and without sufficient legal knowledge, LLMs may struggle to generate accurate reasoning paths. For example,
LMs often fail to distinguish between filing time and trial time or to interpret the specific meanings of different parts of a case number, all of which are crucial for arriving at the correct answer. Although we have provided a legal knowledge corpora, in practice, we find that LLMs often fail to recognize the necessity of consulting relevant legal knowledge. When handling legal issues, LLMs tend to rely on existing patterns and linguistic knowledge, overlooking the precision and details required in the legal domain.
In the future, legal knowledge should be integrated into LLMs more systematically and in-depth to ensure accuracy and reliability in complex legal contexts.  Moreover, LLMs should also intelligently recognize when to access external legal databases, enabling more flexible and accurate decision-making.

\textbf{Insufficient understanding of legal articles and case law.}
The resolution of some legal issues relies on articles and case law, but LLMs may have significant limitations in this regard.
During reasoning, LLMs often struggle to accurately interpret the scope and logic of legal articles. Even when retrievers successfully identify relevant articles and cases, LLMs may still encounter difficulties in understanding the judicial interpretations and practical applications of these referenced materials, especially in complex legal scenarios.
In the future, to enhance the effectiveness of LLMs in the legal domain, it is essential to strengthen their ability to deeply understand and apply legal articles and cases.

\textbf{Other Error Types}
In addition to the above challenges, LLM agents commonly encounter the following errors on LegalAgentBench: 1) Argument Errors: LLM agents fail to provide the correct argument when invoking tools. 2) Planning Errors: LLM agents generate incorrect planning paths or use inappropriate tools due to hallucinations or insufficient knowledge. 3) Exceeding Length Limitations: The encoded interaction history, observations, and tool usage plans exceed the length limit, preventing the LLM agents from resolving the task. 4) Getting Stuck in Loops: LLM agents may repeatedly attempt to solve the same problem, ultimately reaching the maximum iteration limit. Overall, LegalAgentBench provides a comprehensive and reliable evaluation platform, highlighting that LLM agents still have significant room for improvement in solving complex legal problems.

\section{Conclusion}

In this paper, we introduced LegalAgentBench, a benchmark designed to evaluate the performance of LLM agents in the legal domain. To achieve this, we collected real-world data to construct 17 databases and 37 tools.
Additionally, we proposed a scalable task construction framework that comprises six steps: planning tree creation, path selection, entity verification, question rewriting, answer generation, and human verification. This framework is versatile and can be extended to incorporate new knowledge bases and tools, enabling the construction of tasks with varying types and difficulty levels.
We conducted extensive experiments on LegalAgentBench, providing an in-depth analysis of the strengths, weaknesses, and potential improvement directions for existing models and methods. Looking ahead, we aim to expand LegalAgentBench to support additional languages and legal systems, fostering the development of legal LLMs worldwide.

\clearpage

\bibliography{custom}
\clearpage

\appendix
\section{Discussion}
\label{sec:Discussion}
In this section, we discuss limitations, potential impacts, license, and legal and ethical considerations. 
\subsection{Limitation}
\label{sec:Limitation}
We acknowledge several limitations in this study and aim to address them in future work. First, the current dataset is primarily constructed in Chinese, which limits its applicability in broader multilingual contexts. We plan to release an updated version in future iterations that will support English.  Second, the task design primarily covers statutory law system, and its performance in case law systems requires further exploration. It may not fully capture the diversity of legal knowledge and systems across different countries. In future work, we will enrich the task types, broaden the scope, and incorporate legal scenarios from more countries and regions, thereby enhancing the applicability and comprehensiveness of our research.

\begin{table*}[ht]
\centering
\resizebox{\textwidth}{!}{%
\begin{tabular}{cclc}
\hline
ID & Corpus             & Key                                                                                                                                                                                                                                               & \# Key \\ \hline \hline
\rowcolor{mygray}
1  & CompanyInfo        & \begin{tabular}[c]{@{\ }p{0.8\textwidth}@{\ }} Company Name, Company Abbreviation, English Name, Associated Security, Company Code, Former Abbreviation, Market, Industry, Date of Establishment, Listing Date, Legal Representative, General Manager, Board Secretary, Postal Code, Registered Address, Office Address, Telephone Number, Fax, Official Website, Email, Included Indices, Main Business, Business Scope, Company Profile, Par Value per Share (CNY),  Initial Public Offering (IPO) Price (CNY), Net Proceeds from IPO (CNY), Lead Underwriter for IPO\end{tabular} & \textbf{28}               \\ 
2  & CompanyRegister    & \begin{tabular}[c]{@{\ }p{0.8\textwidth}@{\ }} Company Name, Registration Status, Unified Social Credit Code, Legal Representative, Registered Capital, Date of Establishment, Company Address, Contact Number, Contact Email, Registration Number, Organization Code, Number of Insured Persons, Primary Industry Category, Secondary Industry Category, Tertiary Industry Category, Former Name, Company Profile, Business Scope\end{tabular}                                                                          & \textbf{18}      \\ 
\rowcolor{mygray}
3  & SubCompanyInfo     & \begin{tabular}[c]{@{\ }p{0.8\textwidth}@{\ }} Full Name of the Related Listed Company, Relationship to the Listed Company, Listed Company’s Shareholding Percentage, Listed Company’s Investment Amount, Company Name\end{tabular}                                                                                                                                                            & \textbf{5}      \\ 
4  & LegalDoc           & \begin{tabular}[c]{@{\ }p{0.8\textwidth}@{\ }} Related Company, Title, Case Number, Document Type, Plaintiff, Defendant, Plaintiff's Law Firm, Defendant's Law Firm, Cause of Action, Amount Involved (CNY), Judgment Outcome, Date, File Name\end{tabular}                                                                                                                      & \textbf{13}      \\ 
\rowcolor{mygray}
5  & LegalAbstract      & \begin{tabular}[c]{@{\ }p{0.8\textwidth}@{\ }} File Name, Case Number, Text Summary\end{tabular}                                                                                                                                                                                           & \textbf{3}       \\
6  & CourtInfo          & \begin{tabular}[c]{@{\ }p{0.8\textwidth}@{\ }} Court Name, Court Head, Date of Establishment, Court Address, Court Province, Court City, Court County, Court Contact Number, Court Official Website\end{tabular}                                                                                                                                           & \textbf{9}      \\ 
\rowcolor{mygray}
7  & CourtCode          & \begin{tabular}[c]{@{\ }p{0.8\textwidth}@{\ }} Court Name, Administrative Level, Court Level, Court Code, Zoning Code, Rank\end{tabular}                                                                                                                                                                     & \textbf{6}      \\ 
8  & LawfirmInfo        & \begin{tabular}[c]{@{\ }p{0.8\textwidth}@{\ }} Law Firm Name, Law Firm Unique Code, Law Firm Head, Law Firm Registered Capital, Date of Establishment, Law Firm Address, Law Firm Province, Law Firm City, Law Firm County, Contact Phone Number, Contact Email, Law Firm Profile, Law Firm Registration Authority\end{tabular}                                                                                         & \textbf{13}      \\ 
\rowcolor{mygray}
9  & LawfirmLog         & \begin{tabular}[c]{@{\ }p{0.8\textwidth}@{\ }} Law Firm Name, Business Volume Ranking, Served Listed Companies, Listed Companies with Violations During Reporting Period, Listed Companies Under Investigation During Reporting Period\end{tabular}                                                                                                                                            & \textbf{5}      \\ 
10 & AddrInfo           & \begin{tabular}[c]{@{\ }p{0.8\textwidth}@{\ }} Address, Province, City, County\end{tabular}                                                                                                                                                                                         & \textbf{4}      \\ 
\rowcolor{mygray}
11 & RestrictionCase    & \begin{tabular}[c]{@{\ }p{0.8\textwidth}@{\ }} Restricted High-Consumption Enterprise Name, Case Number, Legal Representative, Applicant, Amount Involved (CNY), Executing Court, Filing Date, Restriction Publication Date\end{tabular}                                                                                                                                             & \textbf{4}      \\ 

12 & FinalizedCase      & \begin{tabular}[c]{@{\ }p{0.8\textwidth}@{\ }} Finalized Company Name, Case Number, Person Subject To Execution, Suspected Applicant for Enforcement, Unfulfilled Amount (CNY), Executor Target (CNY), Executing Court, Filing Date, Finalized Date\end{tabular}                                                                                                                                    & \textbf{8}      \\ 
\rowcolor{mygray}
13 & DishonestyCase     & \begin{tabular}[c]{@{\ }p{0.8\textwidth}@{\ }} Dishonest Executed Company Name, Case Number, Dishonest Executed Person, Suspected Applicant for Enforcement, Amount Involved (CNY), Executing Court, Filing Date, Publication Date\end{tabular}                                                                                                                                          & \textbf{9}      \\ 
14 & AdministrativeCase & \begin{tabular}[c]{@{\ }p{0.8\textwidth}@{\ }} Administrative Penalty Company Name, Case Number, Facts, Penalty Outcome, Penalty Amount (CNY), Penalizing Authority, Penalty Date\end{tabular}                                                                                                                                                         & \textbf{8}      \\ 
\rowcolor{mygray}
15 & LegalKonwledge     &   \begin{tabular}[c]{@{\ }p{0.8\textwidth}@{\ }}    \textbf{EXAMPLE}: Tax Law, Tax Collection and Administration Law, Legal Liability
(1) If a taxpayer fails to pay taxes within the prescribed period or if a withholding agent...  \end{tabular}                                                                                                                                                                                                                                               & \textbf{-}      \\
16 & LegalArticle       &     \begin{tabular}[c]{@{\ }p{0.8\textwidth}@{\ }}   \textbf{EXAMPLE}: Article 1082 of the Civil Code of the People's Republic of China
During the woman’s pregnancy, within one year after childbirth, or within six months 
after termination of pregnancy, the man shall not file for divorce...    \end{tabular}                                                                                                                                                                                                                                               & \textbf{-}      \\ 
\rowcolor{mygray}
17 & LegalCases         &                                                                                                                                  \begin{tabular}[c]{@{\ }p{0.8\textwidth}@{\ }}  \textbf{EXAMPLE}: Application for Recognition of a Civil Judgment by Kaohsiung County Court in Taiwan: Kang vs. Huang. (1) Basic Facts: On June 1, 2004, mainland resident Kang...
\end{tabular}                                                                                                                  & \textbf{-}      \\ \hline
\end{tabular}
}
\caption{More details about the corpus of LegalAgentBench.}
\label{corpus:details}
\end{table*}

\subsection{Broader Impact}
\label{sec:Impact}

LegalAgentBench aims to promote the application of LLM agents in the legal field and help legal professionals better understand and evaluate the performance of LLM agents through a standardized evaluation framework. By providing reliable evaluation methods and construction processes, LegalAgentBench not only facilitates the application of LLM agents in legal tasks but also offers valuable insights for building benchmarks in other vertical domains. The widespread use of LLM agents in the legal industry may influence the way legal professionals work, change how they use technological tools, and drive adjustments in legal education and practice. We are particularly focused on the far-reaching impact of LLM agents on the legal industry, ensuring that their application always adheres to principles of social justice and the rule of law. To guarantee fairness and transparency, the datasets and evaluation methods of LegalAgentBench will undergo rigorous ethical reviews and be subject to broad stakeholder involvement to ensure their impartiality.

It is important to note that LegalAgentBench does not advocate for the complete replacement of legal professionals by LLM agents but seeks to reduce the burden on legal personnel and enhance efficiency through supportive tools. The uniqueness and complexity of legal judgment require rich expertise and human insight, qualities that LLM agents cannot fully replace. Our goal is to provide legal professionals with a standardized evaluation tool to help them make more informed decisions in practical applications, understanding when, where, and how to effectively use LLM agents. However, the evaluation results of LegalAgentBench should be used as a reference only, as applications in real-world legal scenarios still require further in-depth evaluation to ensure the legality and reasonableness of decisions. We believe that LegalAgentBench will contribute to the development of a more just and efficient legal system and promote the responsible application of AI technology in the legal field.

\subsection{License}
\label{sec:License}
In this section, we clarify the copyright and licensing status of LegalAgentBench to ensure that users can utilize this dataset in a legal and compliant manner.

In LegalAgentBench, all tabular databases is sourced from authentic, publicly available resources. We have obtained explicit copyright permissions to incorporate these databases into the benchmark. Additionally, the retrieval corpora are derived from publicly accessible legal materials that comply with relevant legal and ethical standards. These resources are made available in accordance with applicable norms for open access to legal information, ensuring that their inclusion in the benchmark does not raise any legal or ethical concerns. While the copyright for these resources remains with the respective government agencies, they have been publicly released and authorized for public use. Users are expected to adhere to the relevant laws, regulations, and provisions established by the respective government agencies when utilizing this data

The entire dataset is released under the MIT License. If you believe that LegalAgentBench contains content that infringes on your copyrighted work, please feel free to contact us at any time to request its removal.

\subsection{Legal and Ethical Considerations}

\label{sec:Ethical}
The construction and release of LegalAgentBench adhere to strict legal and ethical standards to ensure its responsible use in research and development. 
All data in LegalAgentBench has undergone rigorous privacy screening and anonymization processes. Any personal or sensitive information has been removed to comply with applicable data protection laws and ethical research guidelines, ensuring the dataset can be used without compromising individual privacy or security. To prevent potential harm, the datasets in LegalAgentBench have been carefully curated and filtered to exclude discriminatory, explicit, violent, or offensive content. An ethical review by legal experts further ensures that the benchmark minimizes risks related to security, safety, discrimination, surveillance, deception, harassment, human rights, bias, and fairness. By addressing these considerations, LegalAgentBench aims to provide a legally compliant and ethically sound foundation for advancing AI capabilities in the legal domain, while upholding the principles of fairness, transparency, and social responsibility.

\section{Corpus Details}
\label{sec:Corpus}
Table \ref{corpus:details} provides detailed information about the corpora in LegalAgentBench,  it lists the keys for 14 different tabular databases along with the number of keys per table, ranging from 3 to 28. This variation reflects the diversity of scenarios in real-world applications. Additionally, basic examples are provided for the three retrieval corpus to clarify their content.


\section{Tool Details}
\label{sec:Tool}
Tables \ref{tool:input} and \ref{tool:des} provide detailed information about the tools included in LegalAgentBench. Table \ref{tool:input} outlines the input and output for each tool, while Table \ref{tool:des} describes the basic functionality of each tool. 

For each tool, beyond the basic input information, we also provide optional parameters to enhance flexibility in usage. For instance, the Database Tools include the parameter \textit{Columns}, which specifies the keys of the structured data to be returned. Similarly, the text retriever offers a \textit{Number} parameter that allows users to control the number of documents retrieved.
To enhance understanding, Table \ref{tool:call} provides a concrete example of a tool call.

\begin{table*}[]
\centering
\resizebox{\textwidth}{!}{%
\begin{tabular}{clll}
\hline
ID & \multicolumn{1}{l}{Tool}                          & Input                                & Return                                \\ \hline \hline
\rowcolor{mygray}
1  & get\_company\_info                                 & 

\begin{tabular}[c]{@{}l@{}} Company name or abbr. or code \end{tabular} & CompanyInfo                           \\
2  & get\_company\_register                             & Company name                         & CompanyRegister                       \\
\rowcolor{mygray}
3  & get\_company\_register\_name                       & Unified social credit code           & Company name                          \\
4  & get\_sub\_company\_info                            & Company name                         & SubCompanyInfo                        \\
\rowcolor{mygray}
5  & get\_sub\_company\_info\_list                      & Company name                         & List{[}SubCompanyInfo{]}              \\
6  & get\_legal\_document                               & Case number                          & LegalDoc                              \\
\rowcolor{mygray}
7  & get\_legal\_abstract                               & Case number                          & LegalAbstract                         \\
8  & get\_legal\_document\_company\_list                & Company name                         & List{[}LegalDoc{]}                    \\
\rowcolor{mygray}
9  & get\_legal\_document\_lawfirm\_list                & Law firm name                        & List{[}LegalDoc{]}                    \\
10 & get\_court\_info                                   & Court name                           & CourtInfo                             \\
\rowcolor{mygray}
11 & get\_court\_info\_list                             & Province, city, and county         & List{[}CourtInfo{]}                   \\
12 & get\_court\_code                                   & Court name or code                   & CourtCode                             \\
\rowcolor{mygray}
13 & get\_lawfirm\_info                                 & Law firm name                        & LawfirmInfo                           \\
14 & get\_lawfirm\_info\_list                           & Province, city, and county         & List{[}LawfirmInfo{]}                 \\
\rowcolor{mygray}
15 & get\_lawfirm\_log                                  & Law firm name                        & LawfirmLog                            \\
16 & get\_address\_info                                 & Specific address                     & AddrInfo                              \\
\rowcolor{mygray}

17 & get\_restriction\_case                                   & Case number                          & RestrictionCase                       \\
18 & get\_restriction\_case\_company\_list        & Company name                         & List{[}RestrictionCase{]}             \\
\rowcolor{mygray}
19 & get\_restriction\_case\_court\_list & Court name                           & List{[}RestrictionCase{]}             \\
20 & get\_finalized\_case                                      & Case number                          & FinalizedCase                         \\
\rowcolor{mygray}
21 & get\_finalized\_case\_company\_list                                & Company name                         & List{[}FinalizedCase{]}               \\
22 & get\_finalized\_case\_court\_list                         & Court name                           & List{[}FinalizedCase{]}               \\
\rowcolor{mygray}
23 & get\_dishonesty\_case                                   & Case number                          & DishonestyCase                        \\
24 & get\_dishonesty\_case\_company\_list                             & Company name                         & List{[}DishonestyCase{]}              \\
\rowcolor{mygray}
25 & get\_dishonesty\_case\_court\_list                      & Court name                           & List{[}DishonestyCase{]}              \\
26 & get\_administrative\_case                              & Case number                          & AdministrativeCase                    \\
\rowcolor{mygray}
27 & get\_administrative\_case\_company\_list                        & Company name                         & List{[}AdministrativeCase{]}          \\
28 & get\_administrative\_case\_court\_list                 & Court name                           & List{[}AdministrativeCase{]}          \\
\rowcolor{mygray}
29 & legal\_knowledge\_retriever                                   & Query text                          & Relevant knowledge \\
30 & legal\_article\_retriever                                    & Query text                     & Relevant articles \\
\rowcolor{mygray}
31 & legal\_case\_retriever                                    & 
Query text & Relevant cases \\
32 & get\_sum                                           & List{[}num{]}                        & Summation results                     \\
\rowcolor{mygray}
33 & get\_subtraction                                           & Minuend, subtrahend                   & Subtraction results                     \\
34 & get\_multiplication                                           & List{[}num{]}                        & Multiplication results                     \\
\rowcolor{mygray}
35 & get\_division                                           & Dividend, divisor                       & Division results                     \\
36 & get\_rank                                          & List{[}num{]}                        & Sorting results                       \\ 
\rowcolor{mygray}
37 & finish                                          & -                        & Final results                       \\ \hline
\end{tabular}
}
\caption{The input and output of the tools.}
\label{tool:input}
\end{table*}

\begin{table*}[]
\resizebox{\textwidth}{!}{%
\begin{tabular}{cll}
\hline
ID & Tool                                     & Description                                                                                                                                                                                         \\ \hline \hline
\rowcolor{mygray}
1  & get\_company\_info                       & \begin{tabular}[c]{@{\ }p{0.8\textwidth}@{\ }}Query the corresponding listed company information in the {[}CompanyInfo{]} based on {[}Company Name, Company Abbreviation, or Company Code{]}.\end{tabular} \\

2  & get\_company\_register                   & \begin{tabular}[c]{@{\ }p{0.8\textwidth}@{\ }}   Query the corresponding company registration information  in the {[}CompanyRegister{]}  based on {[}Company Name{]}.\end{tabular}                                 \\
\rowcolor{mygray}
3  & get\_company\_register\_name             & \begin{tabular}[c]{@{\ }p{0.8\textwidth}@{\ }}   Query the corresponding company name in the {[}CompanyRegister{]}   based on {[}Unified Social Credit Code{]}.\end{tabular}                                       \\
4  & get\_sub\_company\_info                  & \begin{tabular}[c]{@{\ }p{0.8\textwidth}@{\ }}   Query the corresponding parent company and investment information  in the {[}SubCompanyInfo{]}  based on {[}Company Name{]}.\end{tabular}                         \\
\rowcolor{mygray}
5  & get\_sub\_company\_info\_list            & \begin{tabular}[c]{@{\ }p{0.8\textwidth}@{\ }}   Query all subsidiary companies invested by the parent company  in the {[}SubCompanyInfo{]}  based on {[}Company Name{]}.\end{tabular}                             \\
6  & get\_legal\_document                     & \begin{tabular}[c]{@{\ }p{0.8\textwidth}@{\ }}   Query the corresponding judgment document information  in the {[}LegalDoc{]}  based on {[}Case Number{]}.\end{tabular}                                            \\
\rowcolor{mygray}
7  & get\_legal\_abstract                     & \begin{tabular}[c]{@{\ }p{0.8\textwidth}@{\ }}   Query the text summary of the case in the {[}LegalAbstract{]} based on {[}Case Number{]}.\end{tabular}                                                          \\
8  & get\_legal\_document\_company\_list      & \begin{tabular}[c]{@{\ }p{0.8\textwidth}@{\ }}   Query the corresponding judgment document information in the {[}LegalDoc{]}  based on {[}Company Name{]}.\end{tabular}                                           \\
\rowcolor{mygray}
9  & get\_legal\_document\_lawfirm\_list      & \begin{tabular}[c]{@{\ }p{0.8\textwidth}@{\ }}   Query the corresponding judgment document information in the {[}LegalDoc{]}  based on {[}Law Firm Name{]}.\end{tabular}                                          \\
10 & get\_court\_info                         & \begin{tabular}[c]{@{\ }p{0.8\textwidth}@{\ }}   Query the relevant court information in the {[}CourtInfo{]}  based on {[}Court Name{]}.\end{tabular}                                                             \\
\rowcolor{mygray}
11 & get\_court\_info\_list                   & \begin{tabular}[c]{@{\ }p{0.8\textwidth}@{\ }}   Query the relevant court information in the {[}CourtInfo{]}  based on  {[}Province, City, County{]}.\end{tabular}                                                 \\
12 & get\_court\_code                         & \begin{tabular}[c]{@{\ }p{0.8\textwidth}@{\ }}   Query the relevant court information in the {[}CourtCode{]}  based on  {[}Court Name or Court Code{]}.\end{tabular}                                               \\
\rowcolor{mygray}
13 & get\_lawfirm\_info                       & \begin{tabular}[c]{@{\ }p{0.8\textwidth}@{\ }}   Query the relevant law firm information in the {[}LawfirmInfo{]}  based on  {[}Law Firm Name{]}.\end{tabular}                                                     \\
14 & get\_lawfirm\_info\_list                 & \begin{tabular}[c]{@{\ }p{0.8\textwidth}@{\ }}   Query the relevant law firm information in the {[}LawfirmInfo{]}  based on {[}Province, City, County{]}.\end{tabular}                                            \\
\rowcolor{mygray}
15 & get\_lawfirm\_log                        & \begin{tabular}[c]{@{\ }p{0.8\textwidth}@{\ }}   Query the service records of the law firm in the {[}LawfirmLog{]}   based on {[}Law Firm Name{]}.\end{tabular}                                                    \\
16 & get\_address\_info                       & \begin{tabular}[c]{@{\ }p{0.8\textwidth}@{\ }}   Query the province, city, and county of the address in the {[}AddrInfo{]}  based on {[}Specific Address{]}.\end{tabular}                                         \\
\rowcolor{mygray}
17 & get\_restriction\_case                   & \begin{tabular}[c]{@{\ }p{0.8\textwidth}@{\ }}   Query the relevant high-consumption restriction case information  in the {[}RestrictionCase{]}  based on {[}Case Number{]}.\end{tabular}                          \\
18 & get\_restriction\_case\_company\_list    & \begin{tabular}[c]{@{\ }p{0.8\textwidth}@{\ }}   Query the relevant high-consumption restriction case information  in the {[}RestrictionCase{]}  based on {[}Company Name{]}.\end{tabular}                         \\
\rowcolor{mygray}
19 & get\_restriction\_case\_court\_list      & \begin{tabular}[c]{@{\ }p{0.8\textwidth}@{\ }}   Query the relevant high-consumption restriction case information  in the {[}RestrictionCase{]}  based on {[}Executing Court Name{]}.\end{tabular}                 \\
20 & get\_finalized\_case                     & \begin{tabular}[c]{@{\ }p{0.8\textwidth}@{\ }}   Query the relevant finalized case information in the {[}FinalizedCase{]}   based on {[}Case Number{]}.\end{tabular}                                               \\
\rowcolor{mygray}
21 & get\_finalized\_case\_company\_list      & \begin{tabular}[c]{@{\ }p{0.8\textwidth}@{\ }}   Query the relevant finalized case information in the {[}FinalizedCase{]}   based on {[}Finalized Company Name{]}.\end{tabular}                                    \\
22 & get\_finalized\_case\_court\_list        & \begin{tabular}[c]{@{\ }p{0.8\textwidth}@{\ }}   Query the relevant finalized case information in the {[}FinalizedCase{]}   based on {[}Executing Court Name{]}.\end{tabular}                                      \\
\rowcolor{mygray}
23 & get\_dishonesty\_case                    & \begin{tabular}[c]{@{\ }p{0.8\textwidth}@{\ }}   Query the relevant dishonesty enforcement case information in the  {[}DishonestyCase{]}  based on {[}Case Number{]}.\end{tabular}                                 \\
24 & get\_dishonesty\_case\_company\_list     & \begin{tabular}[c]{@{\ }p{0.8\textwidth}@{\ }}   Query the relevant dishonesty enforcement case information in the  {[}DishonestyCase{]}  based on {[}Company Name{]}.\end{tabular}                                \\
\rowcolor{mygray}
25 & get\_dishonesty\_case\_court\_list       & \begin{tabular}[c]{@{\ }p{0.8\textwidth}@{\ }}   Query the relevant dishonesty enforcement case information in the  {[}DishonestyCase{]}  based on {[}Executing Court Name{]}.\end{tabular}                        \\
26 & get\_administrative\_case                & \begin{tabular}[c]{@{\ }p{0.8\textwidth}@{\ }}   Query the relevant administrative penalty case information in the  {[}AdministrativeCase{]}  based on {[}Case Number{]}.\end{tabular}                             \\
\rowcolor{mygray}
27 & get\_administrative\_case\_company\_list & \begin{tabular}[c]{@{\ }p{0.8\textwidth}@{\ }}   Query the relevant administrative penalty case information in the  {[}AdministrativeCase{]}  based on {[}Company Name{]}.\end{tabular}                            \\
28 & get\_administrative\_case\_court\_list   & \begin{tabular}[c]{@{\ }p{0.8\textwidth}@{\ }}   Query the relevant administrative penalty case information in the  {[}AdministrativeCase{]}  based on {[}Penalizing Authority{]}.\end{tabular}                    \\
\rowcolor{mygray}
29 & legal\_knowledge\_retriever              & Retrieve relevant legal knowledge based on {[}Query text{]}.                                                                                                                                        \\
30 & legal\_article\_retriever                & Retrieve relevant legal articles based on {[}Query text{]}.                                                                                                                                       \\
\rowcolor{mygray}
31 & legal\_case\_retriever                   & Retrieve relevant legal cases based on {[}Query text{]}.                                                                                                                                            \\
32 & get\_sum                                 & Perform {[}Summation{]} based on {[}List{[}num{]}{]}.                                                                                                                                               \\
\rowcolor{mygray}
33 & get\_subtraction                         & Perform {[}Subtraction{]} based on {[}Minuend, Subtrahend{]}.                                                                                                                                       \\
34 & get\_multiplication                      & Perform {[}Multiplication{]} based on {[}List{[}num{]}{]}.                                                                                                                                          \\
\rowcolor{mygray}
35 & get\_division                            & Perform {[}Division{]} based on {[}Dividend, Divisor{]}.                                                                                                                                            \\
36 & get\_rank                                & Perform {[}Sorting{]} based on {[}List{[}num{]}{]}.                                                                                                                                                 \\
\rowcolor{mygray}
37 & finish                                   & Summarize existing information to generate an answer.                                                                                                                                               \\ \hline
\end{tabular}%
}
\caption{The descriptions of the tool functions.}
\label{tool:des}
\end{table*}

\begin{table*}[]
\resizebox{\textwidth}{!}{%
\begin{tabular}{l}
\hline
\begin{tabular}[c]{@{\ }p{\textwidth}@{\ }} \textbf{Example}: Use get\_restriction\_case\_company\_list to query the amounts involved in all cases participated in by \textbf{Jiangsu Yanning New Material Technology Development Co., Ltd}.\end{tabular}               \\ \hline
\begin{tabular}[c]{@{\ }p{\textwidth}@{\ }}\textbf{Call Tool}: \\ \textbf{get\_restriction\_case\_company\_list}(\\             
\hspace{2em} \textbf{identifier}="Jiangsu Yanning New Material Technology Development Co., Ltd.", \\           
\hspace{2em} \textbf{columns}={[}"Amount Involved (CNY)"{]})\end{tabular} \\ \hline
\begin{tabular}[c]{@{\ }p{\textwidth}@{\ }}\textbf{Return}: \\ {[}\{`Amount Involved (CNY)': 686550\}, \{`Amount Involved (CNY)': 385353\}, \{`Amount Involved (CNY)': 17875\}, \{`Amount Involved (CNY)': 2456446\}{]}\end{tabular}                          \\ \hline
\end{tabular}%
}

\caption{A concrete example of a tool call. There are four cases related to the specified company. Since the \textit{columns} parameter specifies the Amount Involved (CNY) attribute, each dictionary in the returned list contains only the Amount Involved (CNY) key. If \textit{}{columns}=[], all keys in the corresponding table will be returned.}
\label{tool:call}
\end{table*}

\section{Task Details}
\label{sec:Task}
Tables \ref{multi:hop} and \ref{writing} provide concrete examples of the tasks included in LegalAgentBench. Table \ref{multi:hop} illustrates a \textbf{multi-hop task} along with a human-provided solution path, showcasing the reasoning steps involved. Table \ref{writing} presents an example of the \textbf{writing task} from the dataset, where the task requires not only querying and retrieving various pieces of information but also processing and organizing the information into a formatted legal defense document. This example highlights the challenges faced by LLM agents when applied to the legal domain, demonstrating the need for advanced reasoning, information synthesis, and adherence to domain-specific requirements.

\begin{table*}[]
\resizebox{\textwidth}{!}{%
\begin{tabular}{l}
\hline
\textbf{Task}: What is the total amount of restricted high consumption for 91320115773957541H?                                                                                                                        \\ \hline
\begin{tabular}[c]{@{\ }p{\textwidth}@{\ }}   \textbf{Step}:\\ 
\textbf{Firstly}, since the provided Unified Social Credit Code "91320115773957541H" cannot be directly used to query the total amount of restricted high consumption, we use the \textbf{get\_company\_register\_name} tool to obtain the company name associated with this code.\\
\textbf{Result of Action: }\\
The query reveals that the company name is Jiangsu Yanning New Material Technology Development Co., Ltd.\\ \\
\textbf{Secondly}, we use the \textbf{get\_restriction\_case\_company\_list} tool to query all restricted high consumption cases involving this company, based on the company name retrieved in the previous step. \\
\textbf{Result of Action: }\\
The query returns the amounts involved in all cases as follows: {[}\{`Amount Involved (CNY)': 686550\}, \{`Amount Involved (CNY)': 385353\}, \{`Amount Involved (CNY)': 17875\}, \{`Amount Involved (CNY)': 2456446\}{]}.\\ \\
\textbf{Thirdly}, we use the \textbf{get\_sum} tool to calculate the total of the amounts obtained in the previous step. \\
\textbf{Result of Action: }\\The calculation yields a total of 3546224 CNY.\\ \\
\textbf{Finally}, we summarize the information and answer the original question.\end{tabular} \\ \hline
\textbf{Answer}: The total amount of restricted high consumption for 91320115773957541H is 3546224 CNY.                                                                                                 \\ \hline
\end{tabular}%
}
\caption{An example of the \textbf{multi-hop task.}}
\label{multi:hop}
\end{table*}

\begin{table*}[]
\resizebox{\textwidth}{!}{%
\begin{tabular}{l}
\hline
\begin{tabular}[c]{@{\ }p{\textwidth}@{\ }}
\textbf{Task}: PersonA has filed a lawsuit against CompanyX. The specific content of the complaint is as follows: {[}Complaint Content{]}. CompanyX has engaged Law Firm A for legal representation. As the head of Law Firm A, please draft a defense statement in response to the complaint based on the specified defense statement format: {[}Defense Statement Format{]}.     \end{tabular} \\ \hline
\begin{tabular}[c]{@{\ }p{\textwidth}@{\ }}   \textbf{Step}:\\ 
\textbf{Firstly}, we need to retrieve the address, legal representative, Unified Social Credit Code, and contact number of CompanyX. \\
Call Tool: get\_company\_register\\\\
\textbf{Secondly}, we need to retrieve the head and contact number of Law Firm A. \\
Call Tool: get\_court\_info\\\\
\textbf{Thirdly}, we need to retrieve relevant legal knowledge, legal article, and legal case \\
Call Tools: legal\_knowledge\_retriever, legal\_article\_retriever, legal\_case\_retriever\\\\
\textbf{Finally}, we generate the answer following the specified defense statement format.
\end{tabular} \\ \hline
\begin{tabular}[c]{@{\ }p{\textwidth}@{\ }}
\textbf{Answer}: \\
Defendant: {[}Company Name{]}, {[}Address{]}, {[}Legal Representative{]}, {[}Unified Social Credit Code{]}, {[}Contact Number{]}\\
Authorized Litigation Representative: {[}Law Firm Name{]}, {[}Law Firm Head{]}, {[}Law Firm Contact Number{]}\\
Plaintiff: {[}Name{]} \\
{[}Specific Content of the Defense Statement{]}                                                                \end{tabular} \\\hline
\end{tabular}%
}
\caption{An example of the \textbf{writing task.}}
\label{writing}
\end{table*}

\section{More Implementation details}
\label{sec:Implementation}

\subsection{Prompt}

In this section, we present the key prompts used in LegalAgentBench. More details can be found in our code. Table~\ref{prompt:rewrite} shows the prompt template for question rewriting during the task construction process. We used GPT-4 to rewrite the original questions and manually reviewed the rewritten questions for accuracy. Tables \ref{prompt:PS}, \ref{prompt:PE} and \ref{prompt:Re} present the prompts used in different methods.

\subsection{Metrics}

In this section, we provide a detailed explanation of the calculation for success rate and process rate.
Assume there is a dataset $\mathcal{D}$ consisting of $N$ data points, where each data point includes a keyword set $\mathcal{K}_i$ and a model output $\mathcal{O}_i$. The rate $s_i$ for the $i$-th data point is calculated as:

$$
s_i = \frac{|\mathcal{M}_i|}{|\mathcal{K}_i|} 
$$
where $\mathcal{M}_i = \{k \in \mathcal{K}_i |k ~\text{appears 
 in} ~\mathcal{O}_i \}$. The notation $|\cdot|$ represents the number of elements in a set. 
When the keyword set $\mathcal{K}_i = \text{key\_answer}$, $s_i$ represents the success rate.
When $\mathcal{K}_i = \text{key\_answer} \cup  \text{key\_middle}$, $s_i$ represents the progress rate. We report the average of all tasks in the experimental results.

\begin{table*}[]\centering
\resizebox{\textwidth}{!}{%
\begin{tabular}{l}
\hline
\textbf{The prompt used in question rewriting.} \\ \midrule
\begin{tabular}[c]{@{\ }p{\textwidth}@{\ }}
You are an advanced question rewriter. Your task is to rewrite the given questions to make them more relevant to real-life scenarios and sound more natural. Please adhere to the following requirements: \\
1. \textbf{Preserve the Core Inquiry}: Do not change the core content or the focus of the original question. Ensure the user's intent is not misunderstood. \\
2. \textbf{Introduce Misleading Context}: Add potentially misleading context to obscure the true purpose of the query. \\
3. \textbf{Maintain Logical Coherence}: The rewritten question should align with daily usage scenarios, have smooth language, and be logically coherent. \\
4. \textbf{Relate to Legal Needs}: Whenever possible, ensure the question remains tied to legal requirements. \\
5. \textbf{Ensure Clarity}: Retain the critical points of the inquiry so that the answers to the original question and the rewritten question remain consistent.
\end{tabular} \\\hline
\end{tabular}%
}
\caption{The prompt template for question rewriting during the task construction process.}
\label{prompt:rewrite}
\end{table*}

\begin{table*}[]
\resizebox{\textwidth}{!}{%
\begin{tabular}{ll}
\hline
\textbf{Stage} &\textbf{The Prompt used in the Plan-and-Solve method.} \\ \midrule
Plan  & \begin{tabular}[c]{@{\ }p{0.9\textwidth}@{\ }} Solve a Question-Answering Task. Please understand the question and develop a step-by-step plan to solve it.\\ Start your output with the title "Plan:" and follow it with a list of steps. Each step should begin with "Step n:", where n is the current step number (1, 2, 3, ...). \\ The plan should include several distinct steps, and completing these steps sequentially will yield the correct answer. \\ Ensure the plan is sufficiently detailed to complete the task accurately, without skipping any steps or adding unnecessary ones. \\ The final step should always be: "Based on the above steps, please answer the original question." \\ At the end of the plan, include the line "End of Plan." \\ You can use the following tools:\\ \{tools\} \\ Relevant data tables and their fields (any field appearing in the table can be used as a value in the columns parameter):\\ \{table\_used\_prompt\}\\ Note: Your task is to develop the plan and output it as requested. Do not execute the plan!\\ Here are some examples:\\ \{examples\}\\ (Examples End)\\ Question: \{question\}\end{tabular} \\ \hline
Solve & \begin{tabular}[c]{@{\ }p{0.9\textwidth}@{\ }} Given a single-step plan, please output the specific action you intend to execute based on the plan. "Action" is specified using a JSON block, which includes an action key (tool name) and an action\_input key (tool input).\\ Valid values for the action key include: "Final Answer" or \{tool\_names\}.\\  You may use the following tools:\\ \{tools\}\\ Relevant data tables and their available fields (any field in the table can be used as a value in the columns parameter):\\ \{table\_used\_prompt\}\\  Each "Action" can only call one tool at a time. If multiple tools need to be called, split them into separate steps.\\ The output of an "Action" must strictly follow the JSON format below and be parsable by Python's json.loads function:\\ ```json\\ \{\{\\     "action": TOOL\_NAME,\\     "action\_input": INPUT\\ \}\}\\'''\\ Here are some examples:\\ \{examples\}\\ (Examples End)\\ Plan you need to execute: \{plan\}\\ Steps you have already completed: \{scratchpad\}\\ Action:\end{tabular}                                                                                                 \\ \hline
\end{tabular}%
}
\caption{The prompt used in the Plan-and-Solve method.}
\label{prompt:PS}
\end{table*}

\begin{table*}[]
\resizebox{\textwidth}{!}{%
\begin{tabular}{ll}
\hline
\textbf{Stage} &\textbf{The prompt used in the Plan-and-Execute method.} \\ \midrule
Replan & \begin{tabular}[c]{@{\ }p{0.9\textwidth}@{\ }}   
Solve a Question-Answering task. Please understand the question and develop a step-by-step plan to solve it.\\ Start your output with the title "Plan:" and follow it with a list of steps. Each step should begin with "Step n:", where n is the current step number (1, 2, 3, ...). \\ The plan should include several distinct steps, and completing these steps sequentially will yield the correct answer. \\ Ensure the plan is sufficiently detailed to complete the task accurately, without skipping any steps or adding unnecessary ones. \\ The final step should always be: "Based on the above steps, please answer the original question."\\  At the end of the plan, include the line "End of Plan."\\ \\ You may use the following tools: \\\{tools\}\\  Relevant data tables and their fields (any field in the table can be used as a value in the columns parameter):\\ \{table\_used\_prompt\}\\ \\ Note: You are only responsible for creating the plan according to the requirements. Do not execute the plan!\\ \\ Examples:\\ \{examples\}\\ (Examples End)\\ \\ Question: \{question\}\\ Your Original Plan: \{plan\}\\ Steps You Have Already Completed: \{scratchpad\} \\ Please update the plan based on the situation. If additional steps are needed, list the remaining steps strictly following the format requirements. Retain the already completed steps as they are and do not repeat them. \\ Each step should begin with "Step n:", where n is the current step number (1, 2, 3, ...).\\ The final step should always be: "Based on the above steps, please answer the original question."\\ At the end of the plan, include the line "End of Plan."
\end{tabular} \\ \hline
\end{tabular}%
}
\caption{The prompt used in the Plan-and-Execute method.}
\label{prompt:PE}
\end{table*}

\begin{table*}[]\centering
\resizebox{\textwidth}{!}{%
\begin{tabular}{l}
\hline
\textbf{The prompt used in the ReAct method.} \\ \midrule
\begin{tabular}[c]{@{\ }p{\textwidth}@{\ }}Solve a Question-Answering Task. The process involves alternating steps of "Thought", "Action", and "Observation".  \\ 
- "Thought" is used to reason about the next step based on the current situation. Note that you only need to consider the immediate next step.  \\ 
- "Action" refers to specifying the tool to use through a JSON block containing an `action' key (tool name) and an `action\_input' key (tool input).  \\
- Valid values for the `action' key include: `"Final Answer" or \{tool\_names\}.  \\
- You may use the following tools:  \\     \{tools\}  \\  
- Relevant data tables and their available fields (any field in the table can be used as a value in the `columns' parameter) include:  \\     \{table\_used\_prompt\}  \\
- Each "Action" can call only one tool. If multiple tools need to be called, break them into separate steps.  \\
- The output of an "Action" strictly follow the JSON format below and be parsable by Python's `json.loads' function:  
\\     ```json\\     \{\{\\         "action": TOOL\_NAME,\\         "action\_input": INPUT\\     \}\}\\     '''
\\ Examples:  \\ \{examples\}  
\\ (Examples End)
\\ Important Notes:\\ 
- When outputting "Action," the result must strictly follow the JSON format specified above.  
\\ - When outputting "Thought," only reason about the immediate next step, and avoid thinking about multiple steps ahead.
\\ Question: \{question\}  \\ \{scratchpad\}\end{tabular} \\ \hline
\end{tabular}%
}
\caption{The prompt used in the ReAct method.}
\label{prompt:Re}
\end{table*}

\section{More Evaluation Result}
\label{sec:Evaluation}
Tables~\ref{table:key_middle_result} and \ref{table:bertscore} report the progress rate and BertScore of different LLMs and methods on LegalAgentBench.
Table~\ref{table:key_middle_result} highlights the progress rate as a more fine-grained evaluation metric.
We observed that GLM-4-Plus surpasses GPT-4o in progress rate across all tasks. This suggests that, although GLM-4-Plus may not perform as well as GPT-4o in terms of final outcomes, it achieves a higher degree of intermediate completion in many tasks. This provides a more comprehensive and fine-grained perspective for analyzing model performance.
In Table~\ref{table:bertscore}, BERTScore measures the semantic similarity between the model's output and the answer. However, the differences between baselines become smaller, making it challenging to effectively distinguish the performance of different baselines. Given the high accuracy requirements in the legal domain, we recommend using success rate and progress rate to evaluate model performance.

\begin{table*}[t]
\centering
\resizebox{\textwidth}{!}{%
\begin{tabular}{llcccccccr}
\hline
Model                              & Method & 1-hop           & 2-hop           & 3-hop           & 4-hop           & 5-hop           & Writing         & ALL             & Tokens   \\ \hline
\multirow{3}{*}{GLM-4}              & P-S    & 0.7519          & 0.5094          & 0.3502          & 0.3238          & 0.2334          & 0.4989          & 0.4984          & 5,100,468  \\
                                   & P-E    & 0.7477          & 0.5225          & 0.4140          & 0.3080          & 0.2631          & 0.4801          & 0.5121          & 9,849,924  \\
                                   & ReAct  & 0.8967          & 0.7092          & 0.5103          & 0.4191          & 0.2820          & 0.5179          & 0.6395          & 11,920,863 \\ \hline
\multirow{3}{*}{GLM-4-Plus}         & P-S    & 0.8231          & 0.6577          & 0.5336          & 0.4791          & 0.3366          & 0.6367          & 0.6304          & 5,657,827  \\
                                   & P-E    & 0.8673          & 0.6531          & 0.5447          & 0.4762          & 0.3156          & 0.6401          & 0.6416          & 9,422,692  \\
                                   & ReAct  & 0.9323          & 0.7835          & 0.6233          & \textbf{0.5788} & 0.4026          & 0.6273          & \textbf{0.7280} & 11,739,861 \\ \hline
\multirow{3}{*}{LLaMa3.1-8B} & P-S    & 0.3475          & 0.2498          & 0.0999          & 0.0734          & 0.0859          & 0.2631          & 0.2123          & 9,279,701  \\
                                   & P-E    & 0.3719          & 0.2248          & 0.1247          & 0.0934          & 0.1097          & 0.3334          & 0.2260          & 13,649,741 \\
                                   & ReAct  & 0.6177          & 0.1573          & 0.0649          & 0.0468          & 0.0581          & 0.1074          & 0.2369          & 50,661,127 \\ \hline
\multirow{3}{*}{Qwen-max}          & P-S    & 0.8485          & 0.6442          & 0.4579          & 0.3994          & 0.2610          & 0.4557          & 0.5907          & 4,800,345  \\
                                   & P-E    & 0.8979          & 0.7004          & 0.5233          & 0.4141          & 0.2836          & 0.5011          & 0.6384          & 9,884,307  \\
                                   & ReAct  & 0.9229          & \textbf{0.7954}          & 0.5832          & 0.5185          & 0.4908          & 0.5659          & 0.7144          & 11,473,873 \\ \hline
\multirow{3}{*}{Claude-sonnet}     & P-S    & 0.8304          & 0.6987          & 0.5975          & 0.4794          & 0.3195          & 0.6575          & 0.6563          & 7,100,962  \\
                                   & P-E    & 0.8262          & 0.7340          & 0.6458          & 0.5188          & 0.4418          & 0.6776          & 0.6890          & 13,566,119 \\
                                   & ReAct  & 0.8867          & 0.7356          & 0.4850          & 0.4533          & 0.3379          & 0.5504          & 0.6493          & 32,878,858 \\ \hline
\multirow{3}{*}{GPT-3.5}           & P-S    & 0.4994          & 0.2856          & 0.1296          & 0.0905          & 0.0412          & 0.0863          & 0.2558          & 5,007,391  \\
                                   & P-E    & 0.5067          & 0.2731          & 0.1254          & 0.0886          & 0.0496          & 0.0973          & 0.2546          & 9,597,807  \\
                                   & ReAct  & 0.6344          & 0.3117          & 0.1162          & 0.1018          & 0.1102          & 0.0813          & 0.3019          & 11,357,664 \\ \hline
\multirow{3}{*}{GPT-4o-mini}       & P-S    & 0.7010          & 0.5371          & 0.4269          & 0.3193          & 0.1637          & 0.5224          & 0.5039          & 5,482,556  \\
                                   & P-E    & 0.7298          & 0.5548          & 0.4525          & 0.3124          & 0.2032          & 0.5547          & 0.5252          & 10,861,492 \\
                                   & ReAct  & \textbf{0.9354}         & 0.6746          & 0.4834          & 0.3652          & 0.3714          & 0.5014          & 0.6329          & 13,332,418 \\ \hline
\multirow{3}{*}{GPT-4o}             & P-S    & 0.6773          & 0.6137          & 0.4544          & 0.3813          & 0.2878          & 0.6336          & 0.5474          & 4,153,333  \\
                                   & P-E    & 0.7669          & 0.6204          & 0.4772          & 0.3955          & 0.2424          & \textbf{0.6748} & 0.5793          & 7,261,238  \\
                                   & ReAct  & 0.9344 & 0.7937 & \textbf{0.6397} & 0.4833          & \textbf{0.4880} & 0.5123          & 0.7199          & 11,206,957 \\ \hline
\end{tabular}%
}
\caption{The process rate of different baselines on LegalAgentBench. P-S represents the Plan-and-Solve method, and P-E represents the Plan-and-Execute method. The best results are highlighted in bold.}
\label{table:key_middle_result}
\end{table*}

\begin{table*}[]
\centering
\resizebox{\textwidth}{!}{%
\begin{tabular}{llcccccccr}
\hline
Model                              & Method & 1-hop           & 2-hop           & 3-hop           & 4-hop           & 5-hop           & Writing         & ALL             & Tokens   \\ \hline
\multirow{3}{*}{GLM-4}              & P-S    & 0.8511          & 0.7585          & 0.6996          & 0.7289          & 0.7459          & 0.7754          & 0.7678          & 5,100,468  \\
                                   & P-E    & 0.8392          & 0.7389          & 0.7131          & 0.7288          & 0.7270          & 0.7725          & 0.7606          & 9,849,924  \\
                                   & ReAct  & 0.9086          & 0.8088          & 0.7913          & 0.7691          & 0.7597          & 0.7705          & 0.8208          & 11,920,863 \\ \hline
\multirow{3}{*}{GLM-4-Plus}         & P-S    & 0.8998          & 0.8036          & 0.7505          & 0.7726          & 0.7805          & 0.7790          & 0.8113          & 5,657,827  \\
                                   & P-E    & 0.9000          & 0.8168          & 0.7453          & 0.7551          & 0.7530          & 0.7791          & 0.8097          & 9,422,692  \\
                                   & ReAct  & 0.9284          & \textbf{0.8718} & \textbf{0.8254} & \textbf{0.8267} & \textbf{0.8277} & 0.7873          & \textbf{0.8630} & 11,739,861 \\ \hline
\multirow{3}{*}{LLaMa3.1-8B} & P-S    & 0.7427          & 0.6674          & 0.6356          & 0.6241          & 0.6313          & 0.6640          & 0.6727          & 9,279,701  \\
                                   & P-E    & 0.7278          & 0.6542          & 0.6421          & 0.6327          & 0.6413          & 0.7336          & 0.6730          & 13,649,741 \\
                                   & ReAct  & 0.8080          & 0.6944          & 0.6583          & 0.6390          & 0.6672          & 0.5931          & 0.7015          & 50,661,127 \\ \hline
\multirow{3}{*}{Qwen-max}          & P-S    & 0.8581          & 0.7518          & 0.7240          & 0.7296          & 0.7338          & 0.7570          & 0.7708          & 4,800,345  \\
                                   & P-E    & 0.8521          & 0.7664          & 0.7259          & 0.7035          & 0.7300          & 0.7602          & 0.7699          & 9,884,307  \\
                                   & ReAct  & 0.9033          & 0.8315          & 0.8196          & 0.7676          & 0.7886          & 0.7835          & 0.8337          & 11,473,873 \\ \hline
\multirow{3}{*}{Claude-sonnet}     & P-S    & 0.8566          & 0.7841          & 0.7452          & 0.7246          & 0.7416          & 0.7928          & 0.7855          & 7,100,962  \\
                                   & P-E    & 0.8538          & 0.7937          & 0.7441          & 0.7314          & 0.7560          & 0.7913          & 0.7888          & 13,566,119 \\
                                   & ReAct  & 0.8722          & 0.8038          & 0.7770          & 0.7510          & 0.7747          & 0.7679          & 0.8053          & 32,878,858 \\ \hline
\multirow{3}{*}{GPT-3.5}           & P-S    & 0.8329          & 0.7344          & 0.6961          & 0.6856          & 0.6317          & 0.5332          & 0.7262          & 5,007,391  \\
                                   & P-E    & 0.8325          & 0.7302          & 0.7008          & 0.6544          & 0.6201          & 0.5412          & 0.7216          & 9,597,807  \\
                                   & ReAct  & 0.8695          & 0.7618          & 0.7182          & 0.6933          & 0.6459          & 0.5125          & 0.7483          & 11,357,664 \\ \hline
\multirow{3}{*}{GPT-4o-mini}       & P-S    & 0.8945          & 0.7811          & 0.7364          & 0.7498          & 0.7427          & 0.7766          & 0.7954          & 5,482,556  \\
                                   & P-E    & 0.8958          & 0.7884          & 0.7376          & 0.7493          & 0.7372          & 0.7785          & 0.7976          & 10,861,492 \\
                                   & ReAct  & \textbf{0.9303} & 0.8307          & 0.7917          & 0.7911          & 0.7900          & 0.7679          & 0.8373          & 13,332,418 \\ \hline
\multirow{3}{*}{GPT-4o}             & P-S    & 0.8966          & 0.7874          & 0.7432          & 0.7387          & 0.7450          & 0.7864          & 0.7983          & 4,153,333  \\
                                   & P-E    & 0.8955          & 0.7822          & 0.7352          & 0.7407          & 0.7489          & \textbf{0.7964} & 0.7962          & 7,261,238  \\
                                   & ReAct  & 0.9154 & 0.8346 & 0.8199 & 0.7848          & 0.8047 & 0.7792          & 0.8409          & 11,206,957 \\ \hline
\end{tabular}%
}
\caption{The BERT-Score of different baselines on LegalAgentBench. P-S represents the Plan-and-Solve method, and P-E represents the Plan-and-Execute method. The best results are highlighted in bold.}
\label{table:bertscore}
\end{table*}

\section{Guidelines for Expert-Annotation}
\label{sec:Guidelines}
To ensure the quality and reliability of the dataset during its construction, we conducted human verification on LegalAgentBench. To guide the validation process and maintain consistency, we provided the following annotation guidelines:

Validation of Answer Accuracy:
Annotators must independently call the relevant tools to generate correct answers, ensuring that the dataset reflects accurate and reliable outputs.
Incorrect answers should be carefully reviewed and corrected to maintain the integrity of the dataset. This process involves cross-verifying outputs with authoritative sources, re-running tool-based evaluations where necessary, and documenting any adjustments made to ensure transparency and reproducibility.

Validation of Query Rewriting:
Annotators must ensure that rewritten queries preserve the meaning and intent of the original query. Specifically, they must verify that the answers obtained from the original and rewritten queries are identical, thereby ensuring semantic equivalence and logical consistency. In addition, any rewritten query that does not align with practical, everyday usage should be modified to ensure naturalness and usability. 

Verification of Relevant Legal Provisions:
Certain tasks, such as Writing tasks, require citing legal knowledge, articles or cases. Annotators must verify that all references are accurate, up-to-date, and relevant to the task at hand. This ensures the legal soundness of the dataset and enhances its applicability in real-world scenarios.

Handling Doubts and Uncertainties:
If annotators encounter doubts or uncertainties during validation, they are required to consult official documents, legal texts, or terminological glossaries associated with the relevant classification system. Collaboration with legal experts is strongly encouraged to resolve ambiguities and clarify issues, ensuring that the dataset remains precise and unambiguous.

Review and Quality Control:
A robust review mechanism is established to maintain high-quality annotations. Senior annotators regularly cross-check and review the annotations, correcting simple errors and refining complex cases. Each annotation undergoes multiple rounds of manual verification to ensure accuracy and consistency. In cases where disagreements arise among annotators, collaborative discussions are held to reach consensus, with the final decision documented to ensure transparency.

Feedback Mechanism:
To promote continuous improvement, a feedback mechanism is in place, allowing annotators to provide insights and suggestions regarding the annotation guidelines. This iterative refinement ensures that the guidelines remain effective, up-to-date, and aligned with the evolving requirements of the dataset.

Ethical Considerations: Ensure that all annotations are conducted with integrity and impartiality, maintaining high standards of accuracy and fairness. Take proactive measures to avoid any biases or conflicts of interest that could compromise the quality or objectivity of the annotations.

By adhering to these guidelines, we aim to produce a high-quality dataset that is not only accurate and reliable but also capable of supporting meaningful advancements in legal AI applications.

\end{document}